\documentclass{article}
\usepackage{spconf,amsmath,graphicx,hyperref}

\usepackage{hyperref}

\usepackage{booktabs}
\usepackage{multirow}
\usepackage{geometry}
\usepackage{subfigure}

\usepackage{amsmath}
\usepackage{graphicx}
\usepackage{lipsum} 

\usepackage{multirow}

\usepackage{amsmath,amsfonts}

\usepackage[ruled]{algorithm2e}

\usepackage{lipsum} 

\usepackage[utf8]{inputenc} 
\usepackage[T1]{fontenc}    
\usepackage{hyperref}       
\usepackage{url}            
\usepackage{booktabs}       
\usepackage{amsfonts}       
\usepackage{nicefrac}       
\usepackage{microtype}      
\usepackage{xcolor}         

\title{FedRD: Reducing Divergences for Generalized Federated Learning via Heterogeneity-aware Parameter Guidance}
\name{Kaile Wang\textsuperscript{*}, Jiannong Cao\textsuperscript{*}, Yu Yang\textsuperscript{\dag}, Xiaoyin Li\textsuperscript{*}, Mingjin Zhang\textsuperscript{*}\thanks{This work is supported in part by Hong Kong RGC Collaborative Research Fund (No.: C5032-23GF); in part by NSFC/RGC Collaborative Research Scheme (No.: CRS\_PolyU501/23); and in part by the Research Institute for Artificial Intelligence of Things, The Hong Kong Polytechnic University.} \thanks{Code is available at: \href{https://github.com/Kailee-WANG/FedRD }{https://github.com/Kailee-WANG/FedRD}}}
\address{\textsuperscript{*}The Hong Kong Polytechnic University, 
\textsuperscript{\dag}The Education University of Hong Kong}
\begin{document}
\ninept
\maketitle
\begin{abstract}
Heterogeneous federated learning (HFL) aims to ensure effective and privacy-preserving collaboration among different entities. As newly joined clients require significant adjustments and additional training to align with the existing system, the problem of generalizing federated learning models to unseen clients under heterogeneous data has become progressively crucial. Consequently, we highlight two unsolved challenging issues in federated domain generalization: \textit{Optimization Divergence} and \textit{Performance Divergence}. To tackle the above challenges, we propose FedRD, a novel heterogeneity-aware federated learning algorithm that collaboratively utilizes parameter-guided global generalization aggregation and local debiased classification to reduce divergences, aiming to obtain an optimal global model for participating and unseen clients. Extensive experiments on public multi-domain datasets demonstrate that our approach exhibits a substantial performance advantage over competing baselines in addressing this specific problem. 

\end{abstract}
\section{Introduction}
\label{sec:intro}
Despite the success of federated learning (FL) \cite{Fedavg}, the problem of \textit{non-independent and identically distributed} (Non-IID) data, commonly presented in the form of label shift and domain shift, has become crucial.
Existing studies of heterogeneous federated learning (HFL) suffer from performance loss in many practical application scenarios, where model derived from participating clients struggles to maintain its effectiveness when applied to unseen clients with \textit{out-of-distribution }(OOD) data. The dilemma raises such a problem: How can federated learning models be generalized to unseen clients under heterogeneous data while ensuring both privacy and high model performance?

Emerging solutions \cite{nguyen2022fedsr,guo2023out} ignore the influence of imbalanced label distributions among clients. Meanwhile, the differences between client model performance in local training and its impact to global model aggregation are seldom noticed. Motivated by the above observations, we identify two unsolved challenging issues in generalized federated learning on non-IID data.
\textbf{i) Optimization Divergence:} We observe a divisive direction of domain and label shift optimization, namely optimization divergence. The direction of domain-invariant information aggregation in conventional federated domain generalization works may differ from eliminating label shift optimization direction, leading to local and global performance degradation. 
\textbf{ii) Performance Divergence:} The rates of convergence and the upper performance limit diverge between clients with different domains. Assuming that an ideal generalized model is achieved through the global aggregation of client updates, a suboptimal client model can hinder the final objective during each communication round. The global generalized model is more likely to obtain a biased optimization direction, which degrades the generalization ability of the global model.

To this end, we propose FedRD, aiming at obtaining a generalized global model from participating clients that achieves optimal performance on unseen clients through reducing the aforementioned divergences. 
We first study the feature extraction process and propose a debiased classifier that mitigates class imbalance through parameter-guided adaptive reweighting. To handle the tradeoffs between privacy and domain generalization, we conceptualize domain knowledge as encoded knowledge within the feature extractor's parametric space, and propose to measure domain discrepancy by specific layer of the updated model. 
Additionally, we introduce a regularization term to compensate for the performance divergence caused by suboptimal client models. 
We conducted extensive experiments on diverse datasets to assess the effectiveness. The experimental results demonstrate that our approach exhibits a substantial performance advantage over competing baselines in addressing this specific problem.

In conclusion, our main contributions are as follows:
i) We study a novel generalized federated learning setting on Non-IID data, identifying two unsolved challenging issues that degrade the overall performance: optimization divergence and performance divergence.
ii) We explore the knowledge capture process in local training and propose FedRD, which reduces the divergences through parameter-guided adaptive reweighting to restrict imbalanced class updates, and a heterogeneity-aware global aggregation strategy to enhance generalization.
iii) Extensive experiments on public multi-domain datasets demonstrate that our approach exhibits a substantial performance advantage over competing baselines.

\begin{figure*} 
  \centering
  \includegraphics[width=0.8\linewidth]{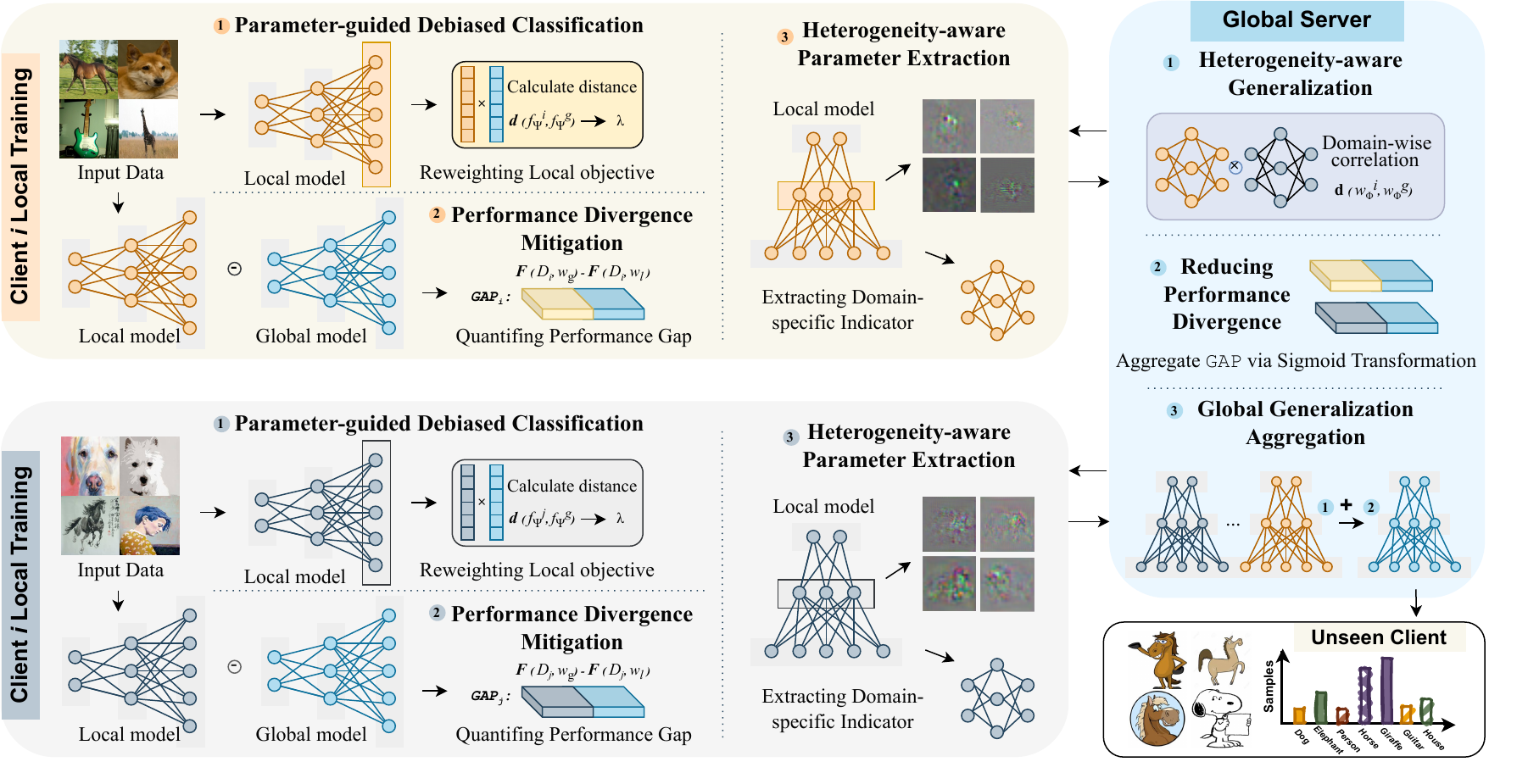}
  \caption{ Overview of FedRD framework. Let $C$ represents the number of classes within the client datasets, the local model, denoted as $f_i(w)=f_{\phi}^{(i)}\circ f_{\psi}^{(i)}$, is composed of a feature extractor $f_{\phi}^{(i)}: \mathbb{R}^X \rightarrow \mathbb{R}^D$ and a debiased classifier $f_{\psi}^{(i)}: \mathbb{R}^D \rightarrow \mathbb{R}^C$. 
On the server side, the global generalization aggregation strategy updates the global model $f_g(w)$. 
 }
  \label{fig:framework} 
\end{figure*}

\section{METHODOLOGY}
\label{sec:METHODOLOGY}

\subsection{Problem Statement}

In this paper, we consider a non-IID federated setting where local data distributions differ among clients. Given a set of $N$ clients with $K_i$ data points $S_i =\left\{\left(x_i^j, y_i^j\right)\right\}_{j=1}^{K_i}$, assume there exists domain shift and label shift among client data distribution $\left(x_i, y_i\right) \sim \mathcal{D}_i$. The overall objective function is to learn a model parameterized by $w$ that minimizes the loss over unseen client distribution $\mathcal{D'}$  by optimizing the following expected risk:
$\min _w \mathcal{E}_{\mathcal{D'}}(w) \approx \frac{1}{N} \sum_{i=1}^N \mathcal{F}(\mathcal{D}_i;w)$,
where $\mathcal{F}(\mathcal{D}_i;w):=\mathbb{E}_{\left(x_i, y_i\right) \sim \mathcal{D}_i}\left[\ell\left(f\left(x_i\right), y_i\right)\right]$, $f(x_i)$ is the mapping function from the input to the predicted label, and $\ell$ is the loss function that penalizes the distance of $f(x_i)$ from $y_i$. 
Let client data be represented by $\mathcal{X} \subset\mathbb{R}^D$, while $\mathcal{Y}$ denotes the label space. Data sample $(x,y)\sim \mathcal{D}: x \in \mathcal{X}, y \in \mathcal{Y}$, and $f_\mathcal{D}: \mathcal{X} \to \mathcal{Y}$ is a deterministic labeling function.
While the unseen test client distribution $\mathcal{D'}$ is inaccessible during training, we came up with our solution to the problem.
\begin{figure}[h]
	\centering
	\includegraphics[width=0.85\linewidth]{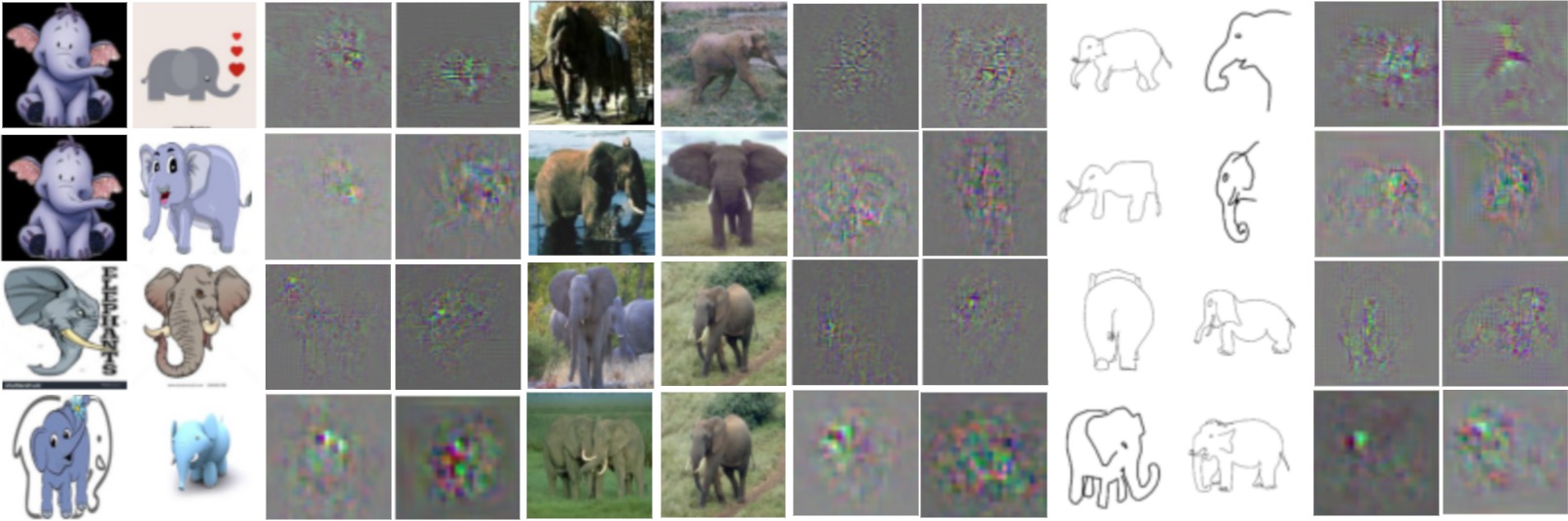}
	\caption{Visualization of images activating each filter in different layers. Best viewed electronically, zoomed in.}
	\label{fig:feature_big}
\vspace{-0.1in}
\end{figure}
\subsection{Feature Extraction}
Referring to the visualization examples of activation figures generated using guided backpropagation \cite{springenberg2015strivingsimplicityconvolutionalnet} to illustrate the desired features of different channels in different layers in Fig. \ref{fig:feature_big}, we made the following observation: Important features of the input picture are captured at different scales. With progressive refinement in feature complexity, deeper layers capture more discriminative, label-related semantic patterns. In lower layers, domain-specific features with more superficial content are comprised \cite{yosinski2015understandingneuralnetworksdeep}. As shown in Fig. \ref{fig:feature_big} and Fig. \ref{fig:feature_small}, where the visualization is selected from the top $10$ image patches activating each filter taken from the local feature extractor, Each row representing one selected layer with three training domains displayed in three columns. 
Consequently, we argue that the lower layer is capable of capturing domain-specific features within single domain local training, and the weights of the lower layer can be regarded as obtained domain knowledge.

\begin{table*}
\label{Domainnetresult}
\centering
\caption{Test Accuracy (\%) on Datasets with Extensive Classes (Mini-DomainNet, OfficeHome) under Heterogeneous Setting.}
\renewcommand{\arraystretch}{1.2}
\resizebox{\textwidth}{!}{
\begin{tabular}{l|cccccc|c|cccc|c}
\toprule
\multirow{2}{*}{\textbf{Methods}} & \multicolumn{7}{c|}{\textbf{Mini-DomainNet}} & \multicolumn{5}{c}{\textbf{OfficeHome}}  \\
\cmidrule(lr){2-8} \cmidrule(lr){9-13}
 & C & I & P & Q & R & S &\textbf{Avg.} & A & C & P & R & \textbf{Avg.}\\
\midrule
FedAvg\cite{Fedavg}   & 50.77(1.25) & 23.46(.89) & 45.85(.32) & 41.32(.47) & 49.90(.65) & 47.74(1.77) & 43.17(.39) & 50.14(.37) & 40.29(.59) & 37.25(1.20) & 49.85(.88) & 44.38(.72) \\
FedProx\cite{li2020federated}  & 52.22(1.18) & 25.63(.87) & 46.47(.25) & 42.21(.22) & 51.89(.75) & 49.01(1.59) & 44.58(1.01) & 50.39(.57) & 42.33(1.04) & 38.97(.45) & 50.86(1.24) & 45.64(.93) \\
Scaffold\cite{pmlr-v119-karimireddy20a} & 52.13(2.28) & 26.25(.94) & 45.80(.44) & 41.76(.62) & 50.36(1.37) & 48.42(.65) & 44.12(.57) & 55.94(.65) & 40.97(1.30) & 41.19(.47) & 55.02(.72) & 48.28(.68) \\
FedSR\cite{nguyen2022fedsr}    & \textbf{56.83}(.34) & \textbf{26.03}(.86) & 48.15(1.45) & 45.32(.67) & 52.19(.54) & 51.44(1.05) & 46.78(.42) & 56.35(.50) & 45.10(.58) & 42.86(.61) & 56.94(.74) & 50.15(.82) \\
FedBN\cite{li2021fedbn}    & 54.65(1.16) & 26.44(.84) & 48.36(.84) & 44.83(.45) & 51.75(.53) & 50.31(.75) & 46.22(.27) & 57.20(.66) & 46.87(.59) & 43.15(.65) & 57.45(.91) & 51.17(.57) \\
FedGA\cite{zhang2023federated}    & 56.31(.29) & 26.95(.88) & \textbf{49.07}(.45) & 46.15(.61) & 54.35(.53) & \textbf{52.84}(.11) & 47.01(.27) & 57.20(.66) & 46.87(.59) & 43.15(.65) & 57.45(.91) & 51.17(.57) \\
FedIIR\cite{guo2023out}   & 55.62(.39) & 27.54(.95) & 48.23(.34) & 46.38(.67) & 52.40(.60) & 50.67(.49) & 45.91(.52) & 57.35(.62) & 46.25(.31) & 44.00(1.62) & \textbf{58.22}(.54) & 51.46(.39) \\
\textbf{FedRD*}   & 55.74(1.14) & \textbf{28.24}(.82) & 48.52(.21) & \textbf{46.66}(.37) & \textbf{54.37}(.26) & 52.41(1.11) & \textbf{47.66}(.16) & \textbf{58.38}(.87) & \textbf{47.14}(.22) & \textbf{44.32}(.95) & 58.17(1.09) & \textbf{52.01}(.81) \\
\bottomrule
\end{tabular}}
\end{table*}

\subsection{Debiased Classifier}
In scenes where training classes are imbalanced between clients, there exist few-shot label sets $M_i$ representing imbalanced class with fewer data points within the $i^{th}$ client training dataset $S_i$. 
For a typical classification task, the cross-entropy loss can be rewritten as follows:
$\mathcal{L}_{local}^{(i)} =  - \sum_{j=1}^{K_i} \left(\sum_{y_j\in {M_i}}^{M_i} y_j \log f_i(x_j)+\sum_{y_j\notin {M_i}}^{Y_i-M_i} y_j \log f_i(x_j) \right)$, 
To restrict the optimization divergence, we propose to address class
imbalance by assigning greater weight to few-shot classes.
We measure the divergence by term $distance$, which is calculated by the Euclidean distance between the global classifier $f_{\psi}^{(g)}$ and local classifier $f_{\psi}^{(i)}$ as follows:
\begin{equation}
\label{eq:distance1}
distance(f_{\psi}^{(g)},f_{\psi}^{(i)}) =  \left[ \sqrt{\sum_{n=1}^{R} (\mathbf{w}_{nm}^{(\psi,g)}- \mathbf{w}_{nm}^{(\psi,i)})^2} \right]_{m=1}^{C}, 
\end{equation}
where $\mathbf{w}_{\psi}^{(g)}$ and $\mathbf{w}_{\psi}^{(i)}$ are the weights of last layer of fully connected layer from global and local classifier, respectively. We then calculate the arithmetic mean of all terms in the resulting vector $\mathbf{v}_{distance}\in \mathbb{R}^{1 \times C}$ denoted as $\lambda$.
Following this, we propose a debiasing local objective by reweighting few-shot classes during training to reduce the effects of class imbalance. We assign class-wise weights $\alpha$ to different classes to conventional local cross entropy loss, where $\alpha = \begin{cases} 
1 + \lambda, & y \in M, \\ 
1, & y \notin M. 
\end{cases}$
Hence the local objective $\mathcal{L}_{local}^{(i)} = -\sum_{j=1}^{K_i} \alpha y_j \log f_i(x_j)$ is rewritten as follows:
\begin{equation}
\label{eq:localloss_new}
\mathcal{L}_{local}^{(i)}  =-\sum_{j=1}^{K_i}(\sum_{y_j\in {M_i}}^{M_i}(1+\lambda) y_j \log f_i(x_j)+\sum_{y_j\notin {M_i}}^{Y_i-M_i} y_j \log f_i(x_j)).
\end{equation}

\begin{figure}[t]
	\centering
	\includegraphics[width=\linewidth]{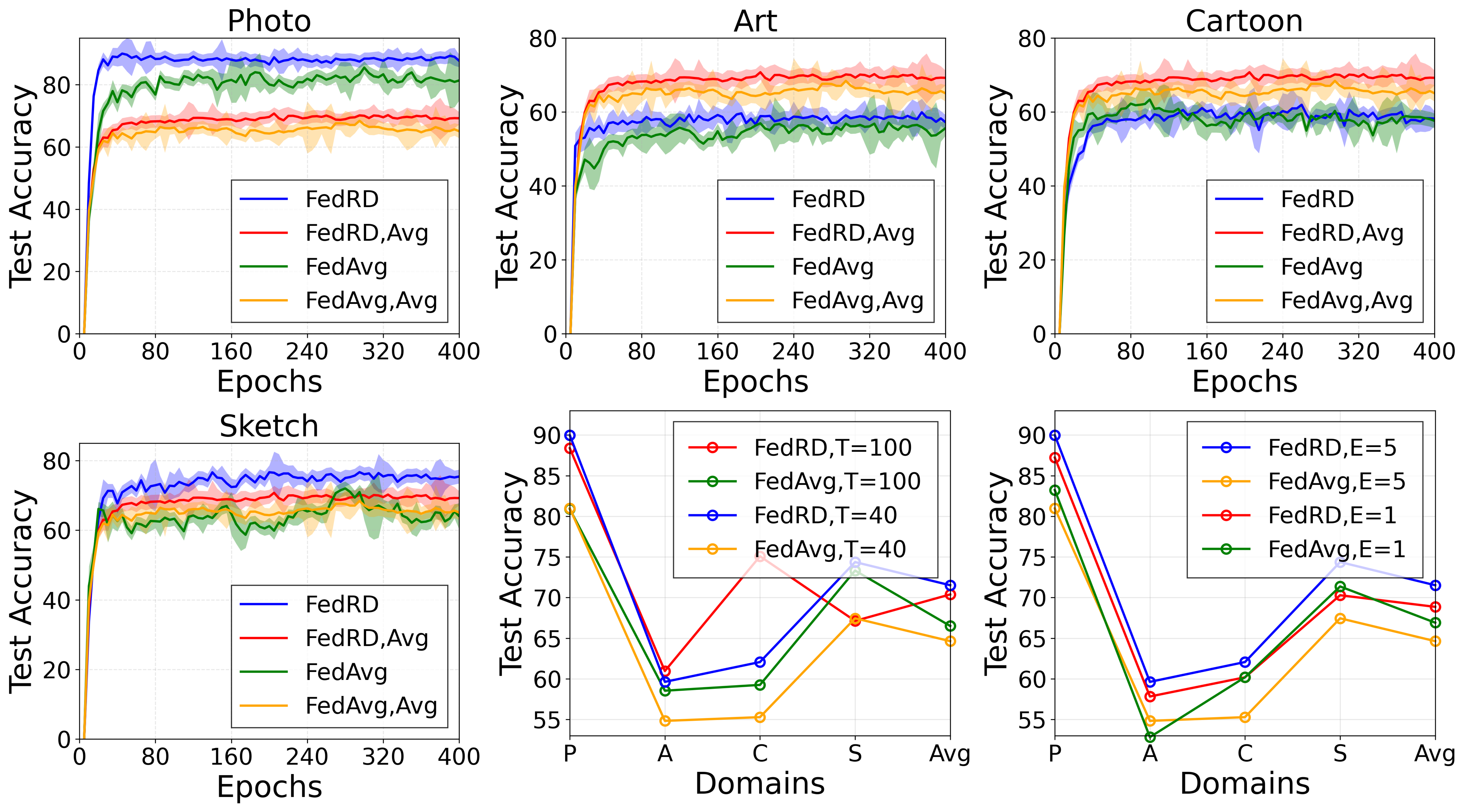}
	\caption{Training performance under different communication rounds.}
	\label{fig:com}
\vspace{-0.2in}
\end{figure}

\subsection{Global Generalization Aggregation}
\label{aggregation}

We conceptualize domain knowledge as encoded knowledge and propose to measure domain discrepancy by specific layer of the updated model. 
Denote $w_{\phi}^{(i)},w_{\phi}^{(g)}\in \mathbb{R}^{I \times O \times H \times W},$ as parameters that exclude the bias of the domain feature extraction layer from the $i^{th}$ client and the global generalized model, respectively.
We define a domain-wise correlation $\mathbf{d_i}$ to measure the heterogeneity between the $i^{th}$ client updated model and global generalized model computed as: $\mathbf{d_i} =  d\langle w_{\phi}^{(i)}, w_{\phi}^{(g)}\rangle =\left\|w_{\phi}^{(i)} - w_{\phi}^{(g)}\right\|_F$.
By calculating domain-wise correlations, the server uncovers latent relationships across domains without compromising client data privacy. 
Additionally, to compensate for the performance divergence caused by suboptimal client models, we introduce a regularization term $\mathcal{GAP}$, measuring the performance gap between the global model and local models on the local data distribution.  This method solves the technical challenge of evaluating performance divergence within a single client and reduces potential bias as follows:
\begin{equation}
\label{eq:gap}
\mathcal{GAP}_i^{(t)} := \left(\mathcal{F}(\mathcal{D}_i;w_g^{(t-1)})-\mathcal{F}(\mathcal{D}_i;w_l^{(t)})\right), 
\end{equation}
where $\mathcal{F}(\mathcal{D}_i;w_g), \mathcal{F}(\mathcal{D}_i;w_l)$ is the objective function of the generalized global model and local model on local data distribution $\mathcal{D}_i$, respectively. 
To mitigate the negative impact of numerical fluctuations on generalization aggregation, we employ sigmoid transformation to scale the $\mathcal{GAP}$. Such that we have $\gamma_i=\frac{1}{e^{-\mathcal{GAP}_i}+1}\in(0,1)$, where $\gamma_i$ persuade a larger weight in the aggregation as a compensation of suboptimum local model.
Consequently, we propose global generalization aggregation, a strategy for enhancing  global model generalization and solving the challenging divergence issues. At $T = t$ round, the global model is updated as follows:
\begin{equation}
\label{globalagg}
f_g^{t+1}(w) = \sum_{i=1}^N\frac{1}{2}\left( \frac{ (1-\beta_i)}{N-1 } +\frac{\gamma_i}{\sum_{i=1}^N \gamma_i}\right)  f_i^{(t)}(w),
\end{equation}
where $\beta_i = \frac{\mathbf{d}_i}{\sum_{i=1}^N \mathbf{d}_i}$.
After the aggregation, the global model is broadcasted to all clients as the initial parameters. 
This approach evaluates performance divergence within a single client and reduces potential bias, enhancing resilience to unseen client data distributions. 

\section{Experiments}
\label{Experiments}
\subsection{Experimental Setup}

\noindent \textit{Datasets.} \space  We evaluate the proposed method on four image datasets: \textbf{PACS} \cite{Li_2017_ICCV}, \textbf{VLCS} \cite{6751316}, \textbf{OfficeHome} \cite{Venkateswara_2017_CVPR} and \textbf{Mini-DomainNet} \cite{peng2019moment}.

\begin{table}[t]
\label{PACSresult}
\centering
\caption{Test Accuracy (\%) on PACS.}
\renewcommand{\arraystretch}{1.2}
\resizebox{\linewidth}{!}{
\begin{tabular}{l|cccc|c}
\toprule
\multirow{2}{*}{\textbf{Methods}} & \multicolumn{5}{c}{\textbf{PACS}} \\
\cmidrule(lr){2-6}
& P & A & C & S & Avg. \\
\midrule

FedAvg\cite{Fedavg}   & 80.98(.25) & 54.83(.92) & 55.29(1.08) & 54.83(.89) & 64.64(.76) \\
FedProx\cite{li2020federated}  & 81.14(.18) & 53.03(.87) & 61.77(1.25) & 71.38(1.32) & 66.83(.15) \\
Scaffold\cite{pmlr-v119-karimireddy20a} & 83.59(.32) & 51.95(.94) & 61.24(1.17) & 70.82(1.28) & 66.90(.72) \\
FedSR\cite{nguyen2022fedsr}    & 85.87(.98) & 53.22(.81) & 61.95(.22) & 73.56(.35) & 68.65(.18) \\
FedBN\cite{li2021fedbn}    & 84.23(1.35) & 52.83(.89) & 62.19(.03) & 72.36(.42) & 67.90(.21) \\
FedGA\cite{zhang2023federated}    & 86.28(.22) & 58.89(.77) & 56.31(.95) & 73.94(1.39) & 69.35(.21) \\
FedIIR\cite{guo2023out}   & 85.26(.41) & 54.58(.33) & \textbf{62.50}(1.28) & 74.45(.46) & 68.53(.35) \\
\textbf{FedRD*} & \textbf{89.98}(.18) & \textbf{59.63}(.72) & 62.07(1.21) & \textbf{74.37}(.42) & \textbf{71.52}(.25) \\
\bottomrule
\end{tabular}}
\end{table}

\setlength{\floatsep}{-0cm}

\begin{table}[t]
\label{VLCSresult}
\centering
\caption{Test Accuracy (\%) on VLCS.}
\renewcommand{\arraystretch}{1.2}
\resizebox{\linewidth}{!}{
\begin{tabular}{l|cccc|c}
\toprule
\multirow{2}{*}{\textbf{Methods}} & \multicolumn{5}{c}{\textbf{VLCS}} \\
\cmidrule(lr){2-6}
& V & L & C & S & Avg. \\
\midrule
FedAvg\cite{Fedavg}   & 76.71(.32) & 48.03(.85) & 56.70(1.14) & 52.97(.97) & 58.60(.88) \\
FedProx\cite{li2020federated}  & 80.51(.24) & 49.99(.93) & 57.12(1.07) & 54.49(.62) & 60.52(.95) \\
Scaffold\cite{pmlr-v119-karimireddy20a} & 77.26(1.35) & 50.39(.86) & 56.35(1.03) & 55.97(.79) & 59.99(.74) \\
FedSR\cite{nguyen2022fedsr}    & 82.31(.41) & \textbf{56.56}(.67) & 63.09(1.19) & 59.33(.95) & 65.35(.05) \\
FedBN\cite{li2021fedbn}    & 82.94(.18) & 49.54(.71) & 62.25(.91) & 59.45(.21) & 63.55(.25) \\
FedGA\cite{zhang2023federated}    & 83.14(.29) & 50.34(1.28) & 59.86(.94) & 64.41(.21) & 64.44(.21) \\
FedIIR\cite{guo2023out}   & 83.25(.13) & 52.83(.73) & 62.38(.91) & 60.34(.96) & 64.95(.32) \\
\textbf{FedRD*} & \textbf{85.37}(.48) & 56.61(.81) & \textbf{63.24}(.17) & \textbf{60.30}(.96) & \textbf{66.38}(.32) \\
\bottomrule
\end{tabular}}
\end{table}


\noindent \textit{Baselines.} \space \space \space We include several federated learning methods as baselines for performance comparison: \textbf{FedAvg} \cite{Fedavg}, \textbf{FedProx} \cite{li2020federated}, \textbf{Scaffold} \cite{pmlr-v119-karimireddy20a}, \textbf{FedSR} \cite{nguyen2022fedsr}, \textbf{FedBN} \cite{li2021fedbn}, \textbf{FedGA} \cite{zhang2023federated} and \textbf{FedIIR} \cite{guo2023out} are chosen for a comprehensive validation.

\begin{figure}[h]
\setlength{\abovecaptionskip}{-0.1cm}
\centering
\begin{subfigure}
[Overall results.]
{\includegraphics[width=0.19\textwidth]{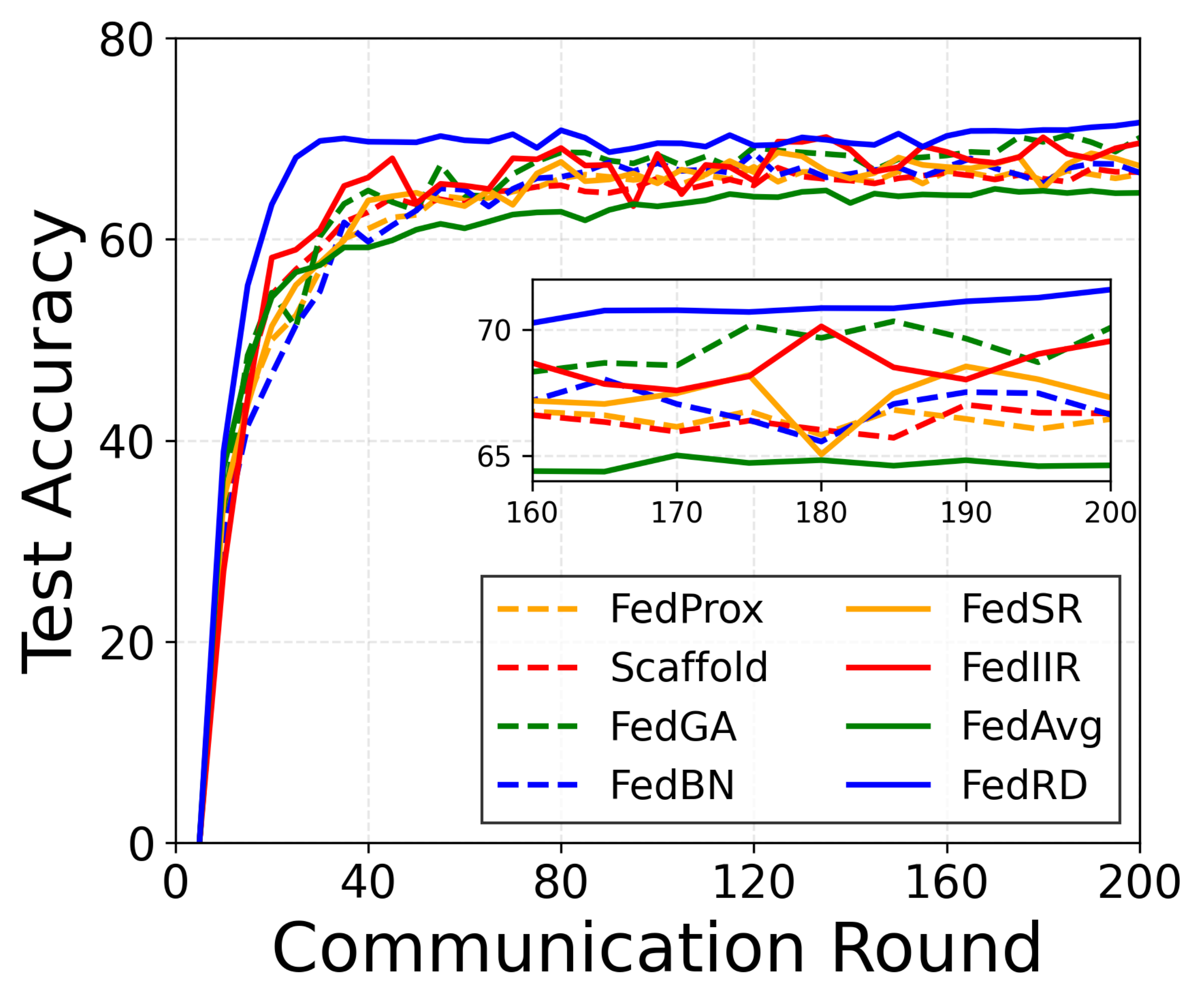}
\label{fig:overall}
}
\end{subfigure}
\quad
\begin{subfigure}
[Feature extraction.
]{\includegraphics[width=0.16\textwidth]{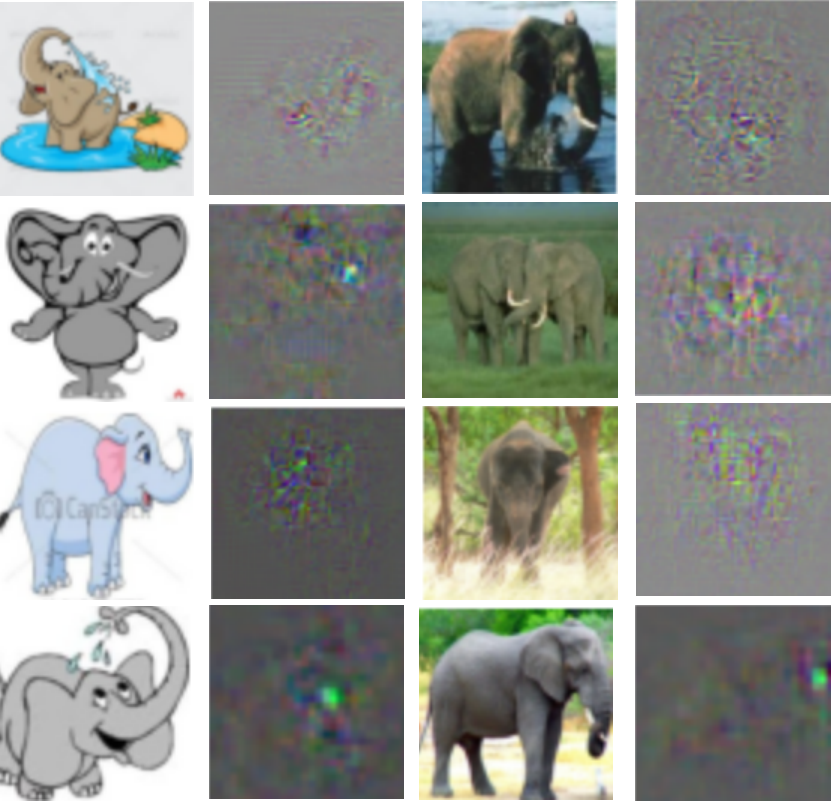}
\label{ablation_all}}
\end{subfigure}
\caption{Training performance and feature extraction analysis.}
\label{fig:feature_small}
\vspace{-0.1in}
\end{figure}

\noindent \textit{Implementation Details.} \space \space \space We conduct leave-one-domain-out evaluation for all datasets. 
For easy comparison with baseline methods, we employ common backbones ResNet18 \cite{he2016deep} and ResNet50 \cite{he2016deep} for feature extraction, and replace the last fully connected layer with 512-dimensional or 2048-dimensional linear layer as classifier.

\subsection{Result Analysis}

We present comparative results on benchmark generalized federated learning experiments on image classification tasks using four publicly available datasets, where FedRD outperforms almost all competing methods by a large margin.
The experimental results are shown in Table \ref{Domainnetresult}, \ref{PACSresult}, \ref{VLCSresult} and Fig. \ref{fig:overall}.
Fig. \ref{fig:feature_big}, \ref{fig:feature_small} are selected samples of top images activating single filter in layer $2, 4, 5$, and $7$ from three different local feature extractors in FedRD.
The first two rows exhibiting activations in layer $2$ and $4$ diverge among domains, while in layer $5$ and $7$ the extractor is more likely to capture specific features and perform comparatively similar extraction as shown on the bottom.

\begin{figure}[t]
\setlength{\abovecaptionskip}{-0.1cm}
\centering
\begin{subfigure}
[Photo.]
{\includegraphics[width=0.22\textwidth]{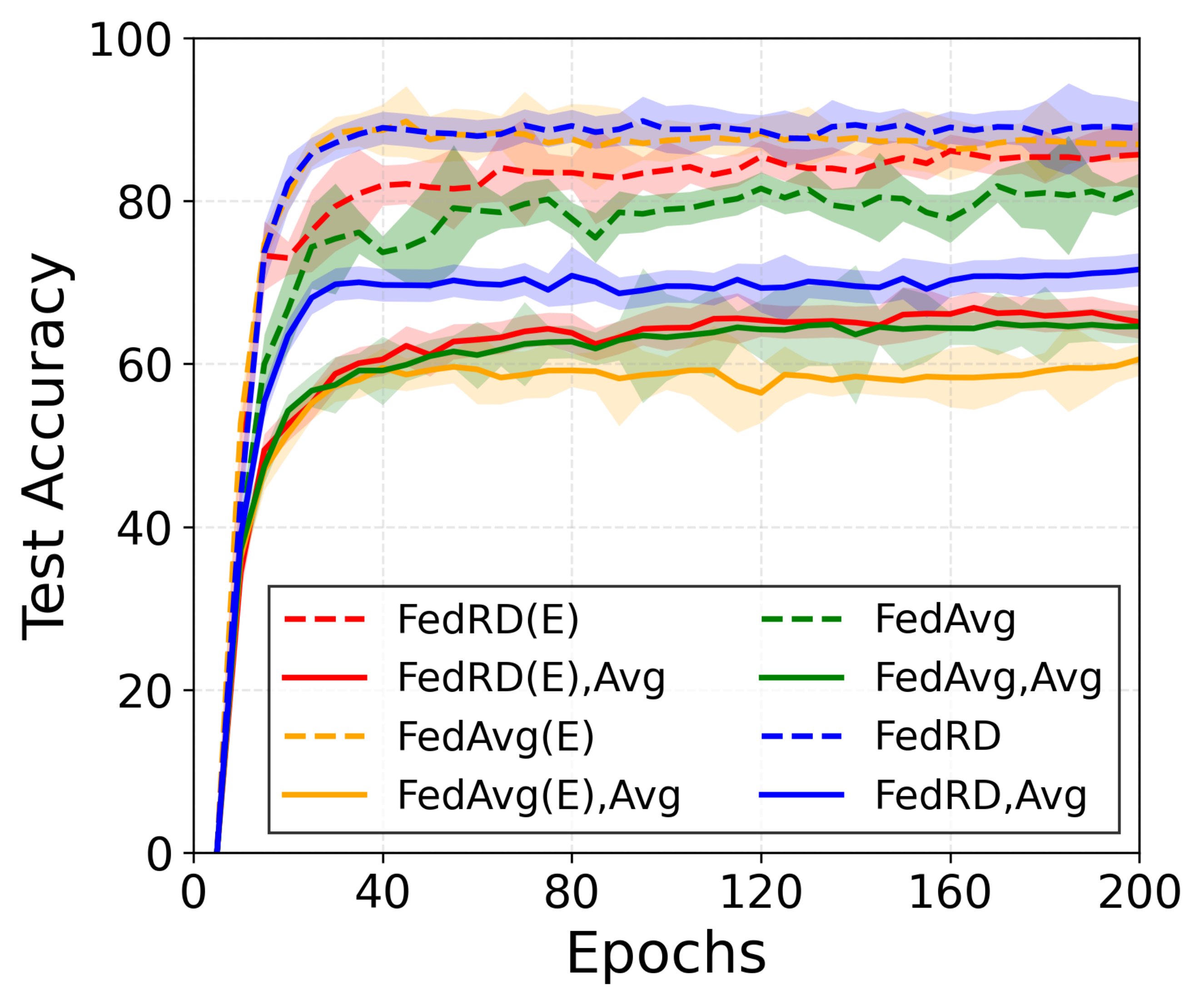}
\label{hetero_p}
}
\end{subfigure}
\hfill
\begin{subfigure}
[Art.
]{\includegraphics[width=0.22\textwidth]{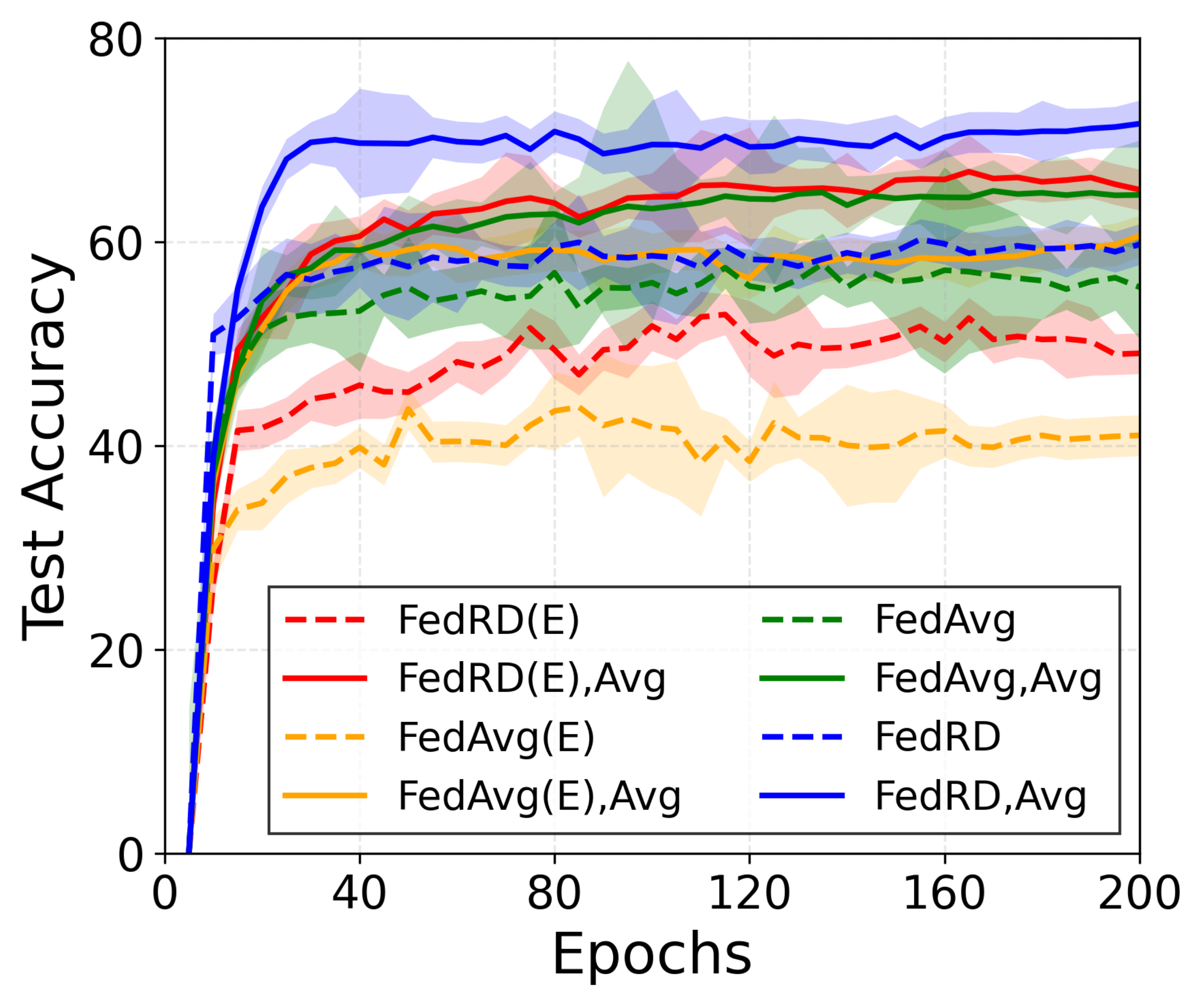}
\label{hetero_a}}
\end{subfigure}
\hfill
\begin{subfigure}
[Cartoon.]
{\includegraphics[width=0.22\textwidth]{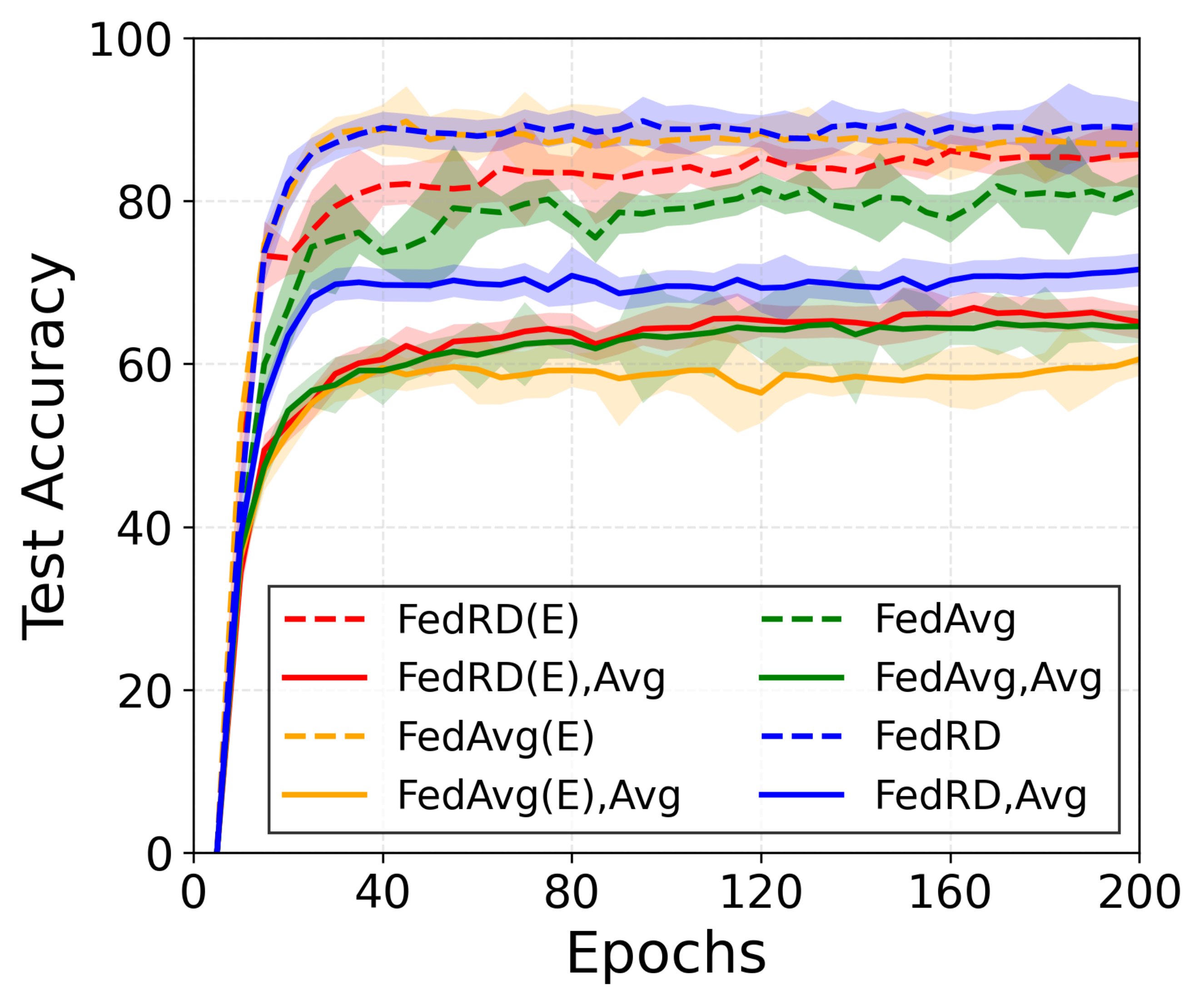}
\label{hetero_c}
}
\end{subfigure}
\hfill
\begin{subfigure}
[Sketch.
]{\includegraphics[width=0.22\textwidth]{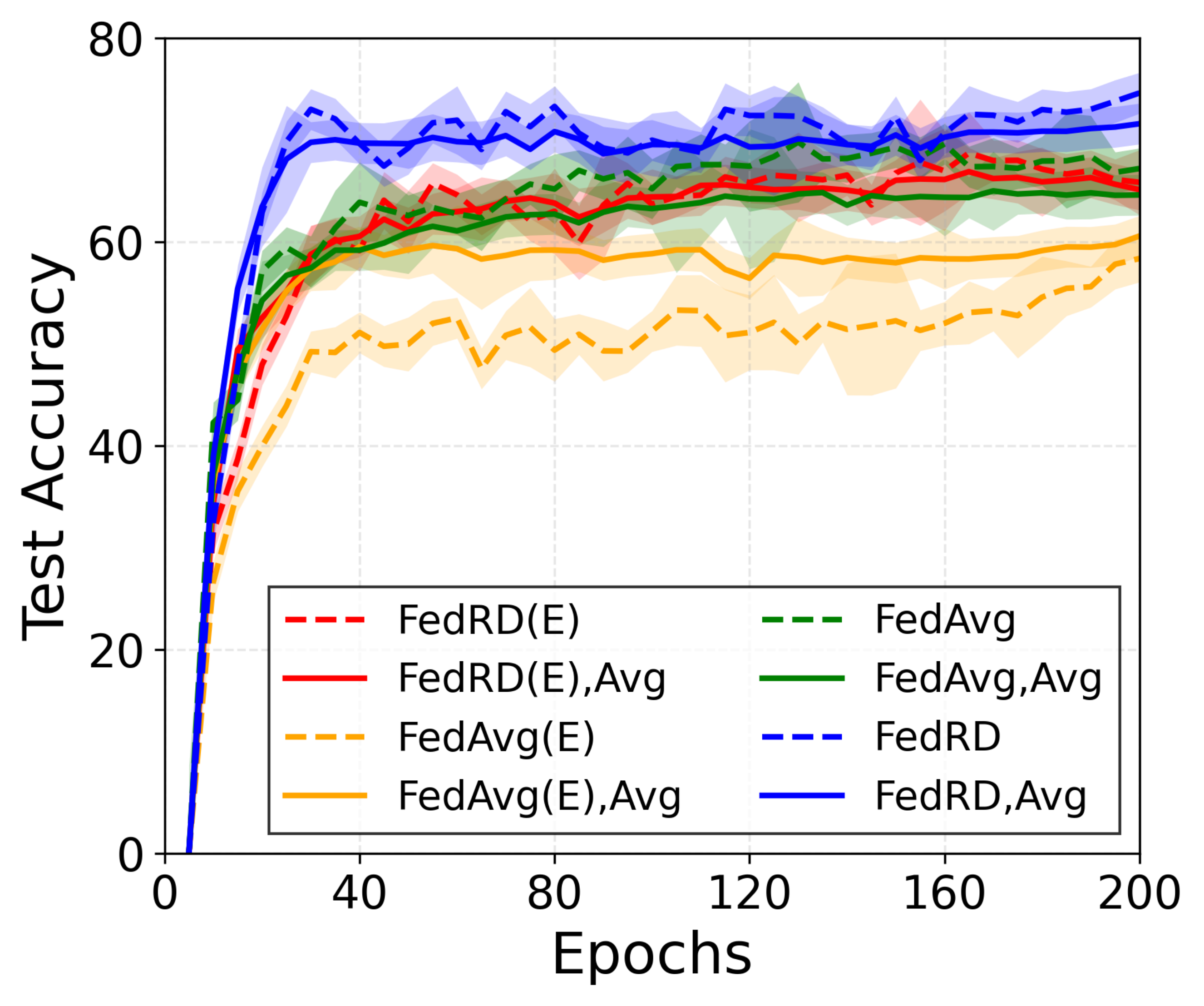}
\label{hetero_s}}
\end{subfigure}
\caption{Training Performance under different heterogeneity.}
\label{fig:hetero}
\end{figure}

\setlength{\textfloatsep}{5pt}

\begin{table}[h]
\label{tab:epoch}
\centering
\caption{Test accuracy (\%) on PACS dataset with different local epochs.}
\renewcommand{\arraystretch}{1.2}
\resizebox{\linewidth}{!}{
\begin{tabular}{l|l|cccc|c}
\toprule
\multirow{2}{*}{\textbf{Epochs}} &\multirow{2}{*}{\textbf{Methods}}& \multicolumn{5}{c}{\textbf{PACS}} \\
\cmidrule(lr){3-7}
 & & P & A & C & S & Avg. \\
\midrule
\multirow{2}{*}{E=1} 
    & FedAvg & 83.23(.31) & 52.83(.25) & 60.20(.71) & 71.37(.88) & 66.91(.25) \\
    & FedRD  & 87.13(.55) & 57.83(1.04) & 61.19(1.23) & 70.29(.71) & 68.80(.80) \\
\midrule
\multirow{2}{*}{E=5} 
    & FedAvg & 80.98(.25) & 54.83(.92) & 55.29(1.08) & 54.83(.89) & 64.64(.76) \\
    & FedRD  & \textbf{89.98}(.18) & \textbf{59.63}(.72) & 62.07(1.21) & \textbf{74.37}(.42) & \textbf{71.52}(.25) \\
\midrule
\multirow{2}{*}{E=10} 
    & FedAvg & 83.11(1.01) & 55.52(.29) & 60.67(.43) & 68.21(.29) & 66.88(.56) \\
    & FedRD  & 88.14(.90) & 58.73(.27) & 61.51(.53) & 75.10(1.17) & 70.87(.48) \\
\bottomrule
\end{tabular}}

\end{table}

\vspace{-1em}
\noindent \textbf{i) Effects of Data Heterogeneity.} 
The results comparing performance of FedRD against baseline methods under varying levels of data heterogeneity as shown in Fig. \ref{fig:hetero} highlight the robustness in handling diverse and imbalanced data distributions.
\textbf{ii) Effects of Local Updating Epochs.}  
The comparison of experiment results with different local updating epochs is shown in Fig. \ref{fig:com} and Table \ref{tab:epoch}, where FedRD demonstrated superior performance across all configurations and achieved its best result at $E=5$.
\textbf{iii) Effects of Communication Round.} We compare the performance in different communication rounds $T = \{40, 100\}$ as shown in Fig.\ref{fig:com}.
FedRD outperforms baseline methods by a large margin in all test domains.
\begin{figure}[t]
\setlength{\abovecaptionskip}{-0.1cm}
\centering
\begin{subfigure}
[Component DC.]
{\includegraphics[width=0.22\textwidth]{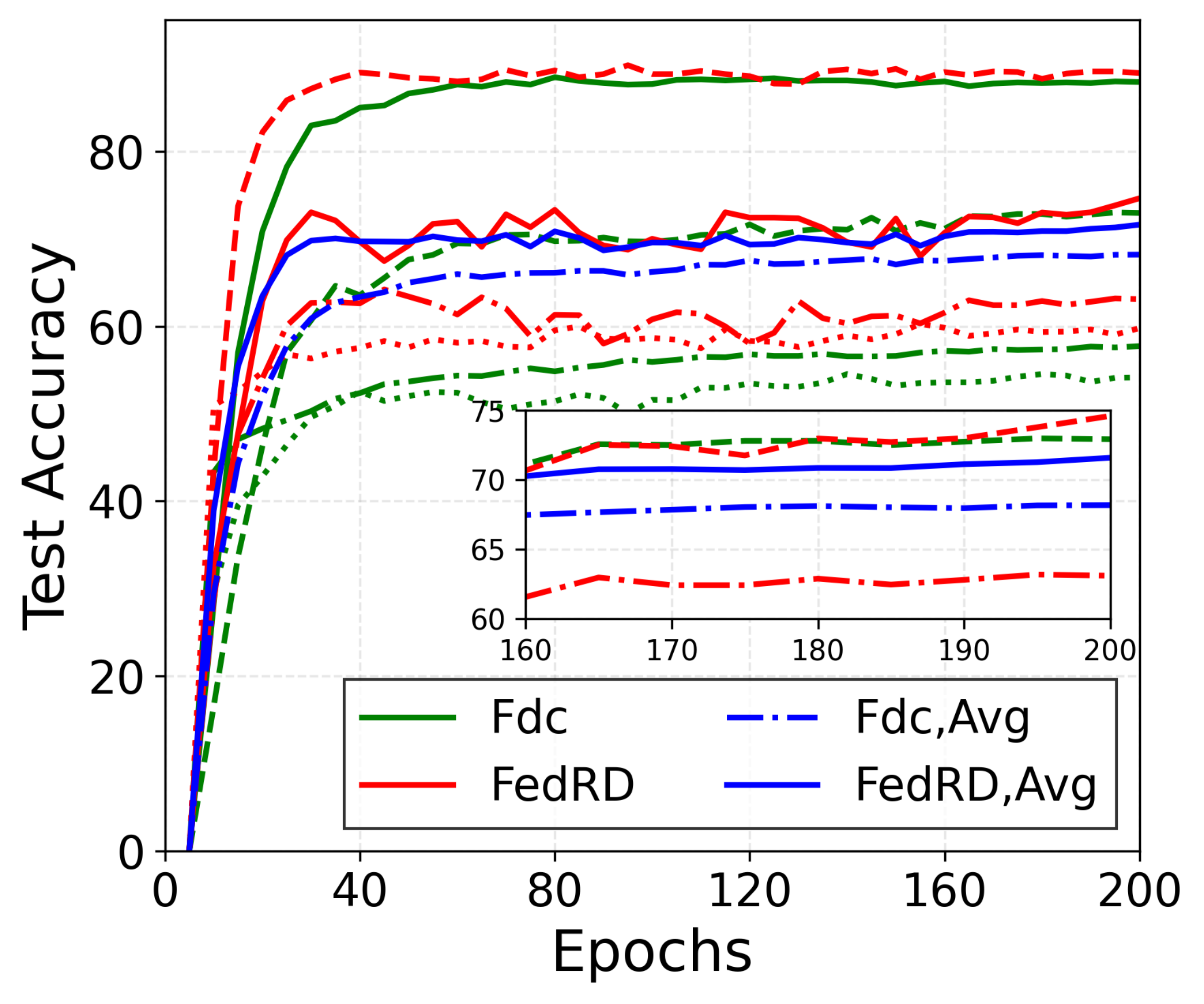}
\label{ablation_dc}
}
\end{subfigure}
\hfill
\begin{subfigure}
[Component GGA.
]{\includegraphics[width=0.22\textwidth]{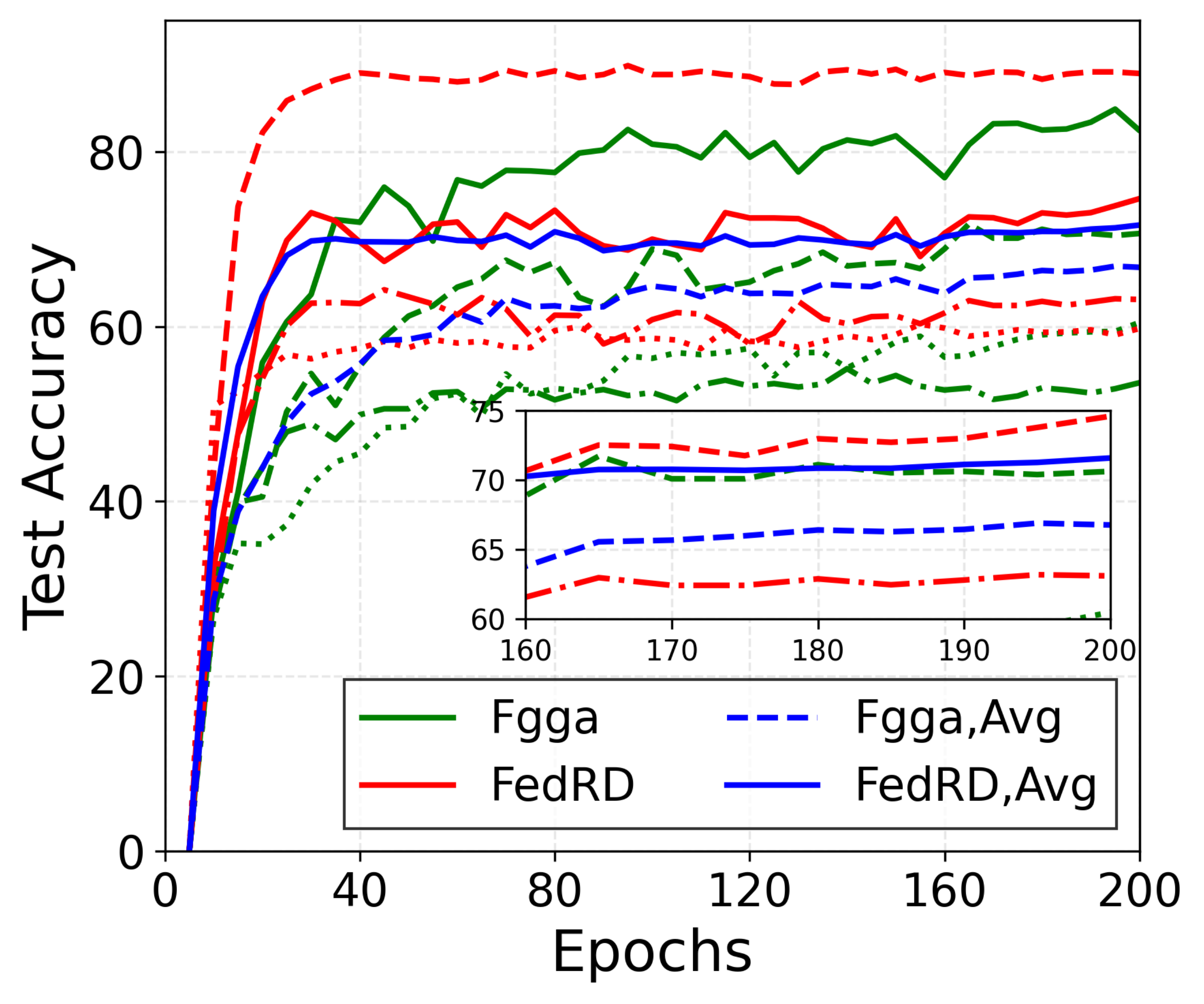}
\label{ablation_gga}}
\end{subfigure}
\hfill
\begin{subfigure}
[Batch size.]
{\includegraphics[width=0.22\textwidth]{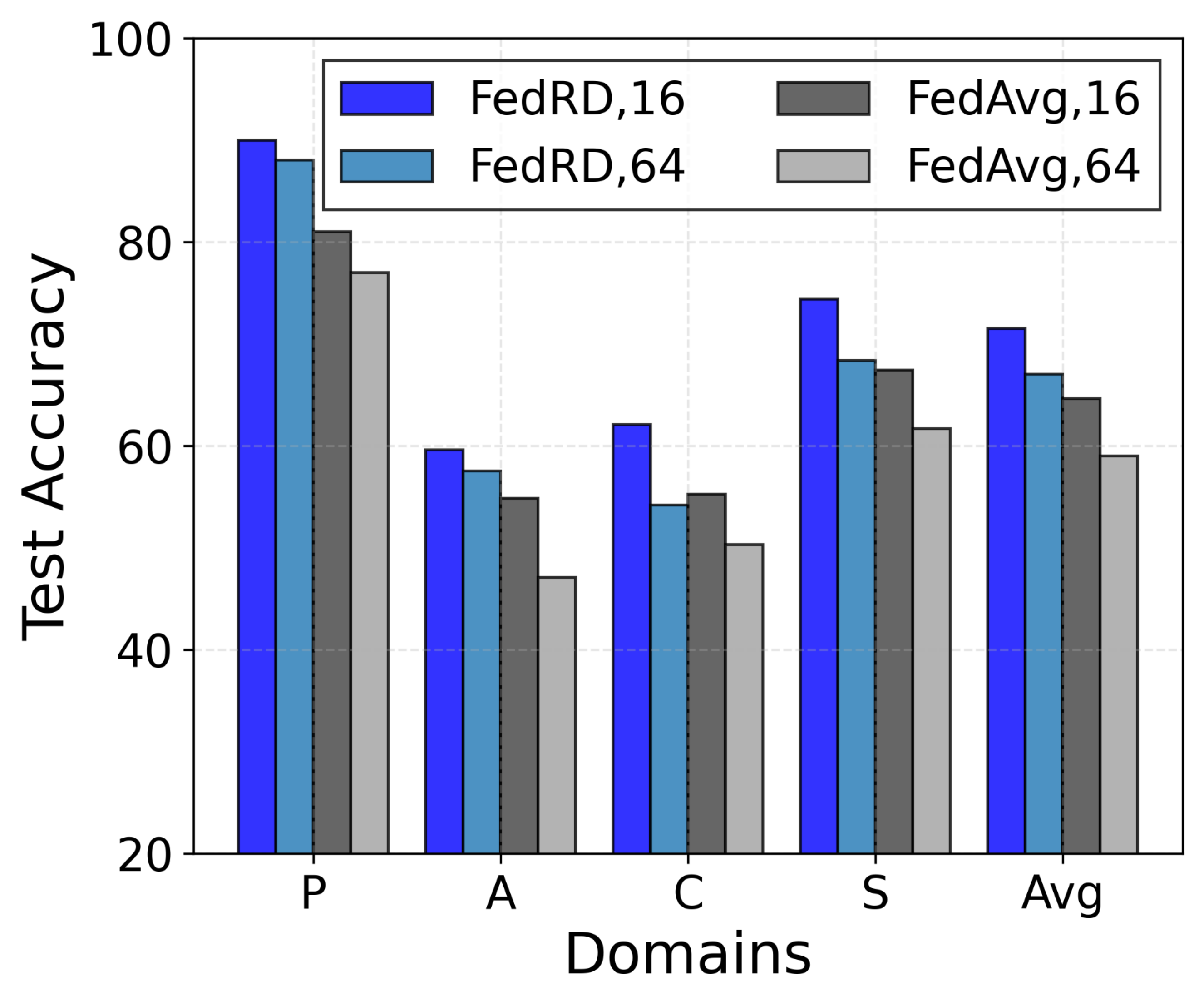}
\label{batchsize}
}
\end{subfigure}
\hfill
\begin{subfigure}
[Learning rate.
]{\includegraphics[width=0.22\textwidth]{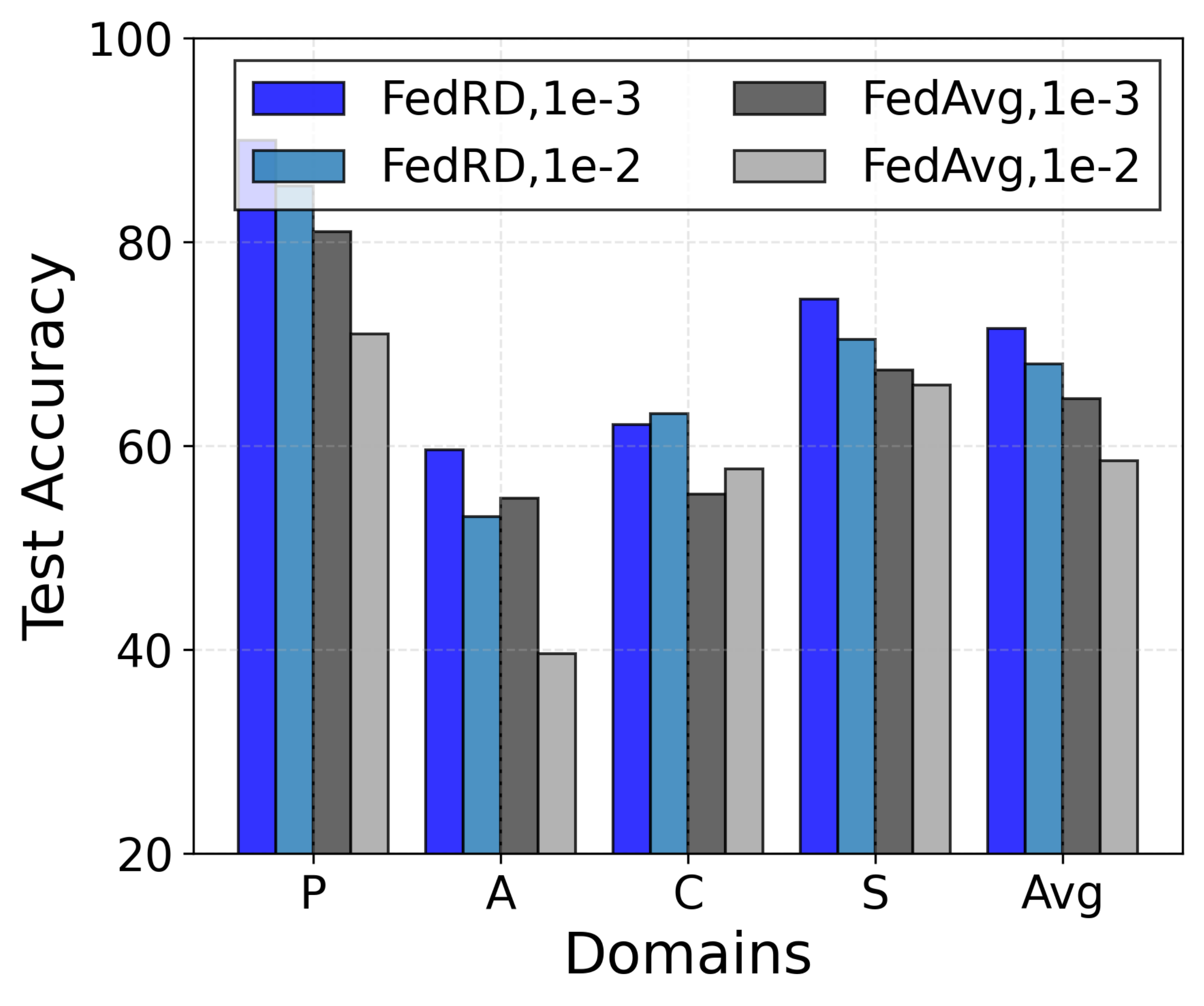}
\label{learningrate}}
\end{subfigure}
\caption{Training Performance under different settings.}
\label{feature_all_abl}
\end{figure}
\textbf{iv) Effects of Local Learning Rate and Batch Size.} 
Effects of learning rate and batch size are explored by assessing performance with different hyper-parameters, as illustrated in Fig.\ref{batchsize}, \ref{learningrate}. The test accuracies of different domains are plotted by bars with different colors.

\setlength{\textfloatsep}{5pt}

\subsection{Ablation Study}
We conducted experiments on the PACS dataset by running our algorithm with different combinations of the two components while maintaining the same experimental setup as the main experiments. Table.\ref{tab:ablation} shows that our method outperforms prior federated learning methods, while the performance in different domains still varies. The learning curve shown in Fig. \ref{ablation_dc}, Fig. \ref{ablation_gga} demonstrates the effectiveness of the proposed method. 
In conclusion, the ablation study demonstrates that the combination of two modules yields the most significant improvements, highlighting the importance of addressing the two divergence issues.

\begin{table}[t]
\label{tab:ablation}
\centering
\caption{Test accuracy (\%) with different
model components.}
\renewcommand{\arraystretch}{1.2}
\resizebox{\linewidth}{!}{
\begin{tabular}{l|cccc|c}
\toprule
\multirow{2}{*}{\textbf{Methods}} & \multicolumn{5}{c}{\textbf{PACS}} \\
\cmidrule(lr){2-6}
& P & A & C & S & Avg. \\
\midrule
\textit{w/o} DC    & 88.02(.63) & 57.57(1.25) & 54.22(.59) & 72.99(1.02) & 68.20(.24) \\
\textit{w/o} GGA   & 83.23(1.30) & 52.83(.78) & 60.19(.88) & 71.37(.16) & 66.91(1.06) \\
\textbf{FedRD}  & \textbf{89.98(.18)} & \textbf{59.63(.72)} & \textbf{62.07(1.21)} & \textbf{74.37(.42)} & \textbf{71.52(.25)} \\ 
\bottomrule
\end{tabular}}
\end{table}


\section{Related Work}

The conventional federated learning FedAvg \cite{Fedavg} usually suffers from data heterogeneity across clients. To solve such problem, some existing works apply fine-tuning \cite{zhou2024every,zhong2023feddar, MLSYS2020_38af8613,chen2023fedcml}, transfer learning \cite{zhang2024eliminating} and sampling \cite{wang2026fedccaclientcentricadaptationdata} to solve the problem, numerous efforts \cite{crawshaw2024federated,ek2025fedalipersonalizedfederatedlearning} made to address the non-IID issue mostly initiate from the perspective of domain and label shift separately.
Despite the success, few studies focuses on more challenging generalized federated learning problems.
Current works attempt with two distinct focuses: improvements to the local training at the client \cite{hu2023generalization,zhang2024improving}, or improvements to the global aggregation \cite{babakniya2024data}.
However, existing explorations mostly target on learning domain-invariant representations for generalization, ignoring the advantages of domain knowledge in global aggregation.

\section{Conclusion}
\label{conclusion}
In summary, we identify a novel challenging issue of divergence in both optimization direction and suboptimum performance in federated generalization problem. We propose FedRD, a novel algorithm that presents a promising approach for generalized federated learning on heterogeneous data. Through the exploration of future research directions such as communication cost and resource constraints, we can further advance the field of generalized federated learning and its practical applications.

\vfill\pagebreak

\bibliographystyle{IEEEbib}
\bibliography{strings,refs}

\end{document}